\documentclass{article}

\usepackage{arxiv}

\usepackage[utf8]{inputenc} % allow utf-8 input
\usepackage[T1]{fontenc}    % use 8-bit T1 fonts
\usepackage{url}            % simple URL typesetting
\usepackage{booktabs}       % professional-quality tables
\usepackage[hidelinks]{hyperref}
\usepackage{amsfonts}       % blackboard math symbols
\usepackage{nicefrac}       % compact symbols for 1/2, etc.
\usepackage{microtype}      % microtypography
\usepackage{cleveref}       % smart cross-referencing
\usepackage{lipsum}         % Can be removed after putting your text content
\usepackage{graphicx}
\usepackage{natbib}
\usepackage{doi}
\usepackage{listings}
\usepackage{xcolor}
\title{KonfAI: A Modular and Fully Configurable Framework for Deep Learning in Medical Imaging}

\date{}

\author{ \href{https://orcid.org/0009-0003-2465-5458}{\includegraphics[scale=0.06]{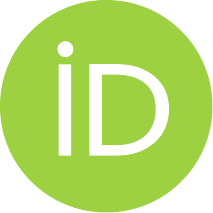}\hspace{1mm}Valentin Boussot} \\
    INSERM, LTSI - UMR 1099\\ 
    University of Rennes\\
    F-35000, France\\
	\texttt{boussot.v@gmail.com} \\
	\And
	\href{https://orcid.org/0000-0001-8840-3944}{\includegraphics[scale=0.06]{orcid.pdf}\hspace{1mm}Jean-Louis Dillenseger} \\
	INSERM, LTSI - UMR 1099\\ 
    University of Rennes\\
    F-35000, France\\
	\texttt{jean-louis.dillenseger@univ-rennes.fr} \\
}

\renewcommand{\headeright}{}
\renewcommand{\undertitle}{}
\hypersetup{
pdftitle={KonfAI: A Modular and Fully Configurable Framework for Deep Learning in Medical Imaging},
pdfsubject={},
pdfauthor={Valentin Boussot, Jean-Louis Dillenseger},
pdfkeywords={Deep Learning, Image registration, Image segmentation, Image synthesis, Medical image analysis},
}

\begin{document}
\maketitle

\begin{abstract}
	KonfAI is a modular, extensible, and fully configurable deep learning framework specifically designed for medical imaging tasks. It enables users to define complete training, inference, and evaluation workflows through structured YAML configuration files, without modifying the underlying code. This declarative approach enhances reproducibility, transparency, and experimental traceability while reducing development time. Beyond the capabilities of standard pipelines, KonfAI provides native abstractions for advanced strategies including patch-based learning, test-time augmentation, model ensembling, and direct access to intermediate feature representations for deep supervision. It also supports complex multi-model training setups such as generative adversarial architectures. Thanks to its modular and extensible architecture, KonfAI can easily accommodate custom models, loss functions, and data processing components. The framework has been successfully applied to segmentation, registration, and image synthesis tasks, and has contributed to top-ranking results in several international medical imaging challenges. KonfAI is open source and available at \href{https://github.com/vboussot/KonfAI}{https://github.com/vboussot/KonfAI}.
\end{abstract}

% keywords can be removed

\section{Introduction}

Deep learning has become central to medical image analysis, yet building complete and maintainable pipelines remains complex and time-consuming. While PyTorch has become the dominant library for deep learning research thanks to its dynamic computation model, Pythonic design, and strong community support~\cite{paszke2019pytorch}, it remains a low-level framework focused on efficient tensor operations and automatic differentiation. These strengths enable rapid prototyping of neural networks, but do not provide high-level abstractions for building complete workflows. In practice, building full workflows still requires significant boilerplate code. Training loops, data preprocessing, inference routines, and evaluation logic are often handcrafted, scattered across scripts, and tightly coupled to specific use cases.

Essential features such as patch-wise training, test-time augmentation (TTA), model ensembling, deep supervision, and multi-output learning remain poorly abstracted in most frameworks. These capabilities are critical in medical imaging, yet researchers frequently re-implement them for each project. This slows development, increases the risk of bugs, and diverts attention from experimental goals. High-level libraries like HuggingFace Transformers~\cite{wolf2020transformers} demonstrate how standardized APIs and reusable components can simplify deep learning workflows. By reducing engineering burden, they make experimentation faster.

The lack of standardization also compromises reproducibility. When experimental settings are embedded in code rather than declared explicitly, it becomes difficult to track changes, compare approaches, or reproduce results reliably. As noted by Pineau et al.~\cite{pineau2021improving}, many reproducibility failures stem from under-specified or poorly structured pipelines.

To address these challenges, we introduce KonfAI: a modular framework designed to simplify the development of flexible and reproducible medical AI pipelines. All aspects of an experiment, including models, data processing, augmentations, loss functions, optimizers, schedulers, post-processing, and more, are explicitly defined through hierarchical YAML configuration files. A well-designed abstraction layer separates experimental intent from implementation details without compromising extensibility.
 
The framework also includes native abstractions for essential capabilities such as test-time augmentation, model ensembling, patch-based reconstruction, and multi-output supervision, making them directly accessible without additional coding. In addition to configuration, KonfAI organizes each experiment within a dedicated working directory, storing logs, predictions, checkpoints, and evaluation results in a consistent and traceable manner. This built-in experiment management system supports reproducible research and streamlines model development, benchmarking, and deployment.

We demonstrate KonfAI’s versatility across segmentation, image registration, and image synthesis, including top-ranking performances in international medical imaging challenges. By combining structured configuration with a modular design, KonfAI enables the creation of robust, scalable, and interpretable pipelines while preserving both flexibility and transparency. Its declarative interface supports rapid experimentation, consistent benchmarking, and accelerated research and development.

\section{Related Work}

The need for reproducible and standardized deep learning workflows is widely recognized in the research community. In recent years, several high-level frameworks have been built as layers on top of PyTorch to address these issues. They range from general-purpose training orchestrators to domain-specific toolkits, each tackling part of the problem but often with trade-offs in flexibility, extensibility, or ease of configuration, and without offering full configurability across the entire pipeline.

\paragraph{General-purpose orchestration.}
PyTorch Lightning~\cite{Falcon_PyTorch_Lightning_2019} abstracts much of the boilerplate involved in training, logging, checkpointing, and device management. It is often combined with \textit{Hydra}~\cite{Yadan2019Hydra}, which offers a structured approach to configuration and improves control over experimental parameters. However, even when used together, they do not fully decouple configuration from code, since key elements of the training logic like loss definitions, data handling, and learning rate scheduling still require manual implementation. In addition, PyTorch Lightning does not provide the high-level abstractions required to support complex, domain-specific workflows commonly encountered in medical imaging.

\paragraph{Medical imaging frameworks.}
Earlier efforts like \textit{NiftyNet}~\cite{gibson2018niftynet} aimed to bring structure and reusability to medical image analysis through configuration files and modular components for tasks such as segmentation and regression. While it contributed to early efforts in reproducibility, it was built on TensorFlow ~\cite{abadi2016tensorflow1}, is no longer maintained, and does not align with the current PyTorch-centered ecosystem. More recently, \textit{MONAI}~\cite{Cardoso_MONAI_An_open-source_2022} has emerged as a PyTorch-based framework specifically developed for medical imaging, offering a wide range of reusable components for models, losses, and data transformations, along with a built-in training engine. However, this engine follows an event-driven design that can be rigid in practice. Extending MONAI to support advanced or non-standard workflows such as multi-output training or deep supervision often requires bypassing or rewriting parts of its internal training logic. In addition, MONAI does not fully separate configuration from code, making it harder to define and reproduce experiments in a purely declarative manner.

\paragraph{Task-specific automation.}
Tools like \textit{nnU-Net}~\cite{isensee2021nnu} and \textit{DeepReg}~\cite{fu2020deepreg} aim to fully automate the segmentation or registration pipeline, from data preprocessing to model training and post-processing. These frameworks are well-suited to their respective domains. nnU-Net is specifically designed for segmentation using U-Net variants, whereas DeepReg targets image registration exclusively. However, their specialization makes them less suitable for broader use cases such as image synthesis, multi-task learning, or the exploration of novel architectures and training strategies.

While existing solutions each address parts of the problem, none provide all of the following: a fully declarative approach to pipeline design, native support for advanced medical imaging workflows, and standardized experiment layouts for reproducibility. KonfAI is designed to fill this gap by combining modularity, transparency, and extensibility, without compromising research flexibility.

\section{Design Principles}

KonfAI is built on design principles that address common challenges in developing deep learning pipelines for medical imaging. These principles aim to improve transparency, reproducibility, and modularity while preserving the flexibility required for research. Built entirely in PyTorch, the framework benefits from its dynamic computation model and extensive ecosystem, making it both familiar to most users and straightforward to extend.

\paragraph{1. Declarative configuration.}
KonfAI specifies every aspect of an experiment, from data and models to training and evaluation strategies, using structured YAML files. These configuration files act both as the full specification of the pipeline and as a complete, self-contained record of the experimental setup. All design choices, including hyperparameters, are made explicit, traceable, and reproducible. 

Beyond basic training parameters such as batch size or number of epochs, each component is identified by its Python class path, and all constructor arguments are defined directly in the configuration. This guarantees full control and transparency over initialization, eliminates hardcoded values, and enables the entire pipeline to be reconstructed from configuration alone.

\paragraph{2. Modularity and controlled extensibility.}
KonfAI is built around modular, interchangeable components that can be configured, replaced, or extended independently. These include models, losses, transformations, schedulers, augmentations, and more. At runtime, all elements are instantiated through a unified registry system grounded in well-defined interfaces. Each module inherits from a shared base class such as \texttt{Network}, \texttt{Loss}, or \texttt{Transform}, ensuring compatibility across the pipeline and enabling seamless integration of new components.

The framework ships with a comprehensive set of built-in modules covering most needs in medical imaging. Custom extensions can be introduced by subclassing the relevant interface and referencing the new module directly in the configuration, without altering the core codebase. Replacing a UNet with a SwinUNet, or switching from Dice loss to a hybrid objective, only requires a single-line modification.

\paragraph{3. Experiment traceability and workspace management.}
Since every aspect of the pipeline is declared in the configuration, KonfAI can fully trace each experiment by storing it in a dedicated workspace containing configuration snapshots, logs, predictions, checkpoints, and evaluation results. This consistent organization makes experiments self-contained, comparable, and auditable, ensuring that they can be reproduced or extended without ambiguity. By preserving both the configuration and the outputs, the workspace serves as a complete record of the experiment, facilitating debugging, benchmarking, and retrospective analysis.

\section{Command-Driven Workflow and Workspace Organization}

KonfAI’s workflow is built around three primary commands: \texttt{TRAIN}, \texttt{PREDICTION}, and \texttt{EVALUATE}. Each command is driven by its own YAML configuration file, \texttt{Config.yml}, \texttt{Prediction.yml}, and \texttt{Evaluation.yml}, which specifies every parameter needed for execution.

The training configuration (\texttt{Config.yml}) defines the model architecture, datasets, loss functions, transformations, augmentations, learning rate schedulers, output heads, and evaluation metrics. The prediction configuration (\texttt{Prediction.yml}) describes how to load the trained model(s), prepare the input data, apply optional strategies such as patch-based inference or test-time augmentation, and write outputs to disk. The evaluation configuration (\texttt{Evaluation.yml}) specifies how to compare predictions against ground truth, including metric definitions, aggregation rules, and output formats.

Executing any of these commands automatically generates or updates a structured workspace organized by model name. For training, the workspace stores checkpoints, TensorBoard logs~\cite{abadi2016tensorflow}, and a snapshot of the configuration used. Prediction adds the generated outputs and related logs, while evaluation compiles the computed metrics into structured JSON files, including both per-case scores and aggregate summary statistics. This progressive accumulation of artifacts provides a complete chronological record of the experiment without manual file management.

An example workspace generated for a model named \texttt{UNet} is shown in Appendix~\ref{Appendix:workspace-structure}, and the corresponding configuration files are provided in Appendix~\ref{appendix:config-files}. These examples illustrate how KonfAI’s configuration-driven design allows the same codebase to be reused across diverse tasks simply by editing YAML.

\section{Core Features}

KonfAI offers advanced capabilities designed to meet the specific challenges of medical image analysis. These include patch-based learning in 2D, 2.5D, and 3D, test-time augmentation, model ensembling, deep supervision, and flexible prediction strategies.

In addition to conventional workflows, the framework supports complex training setups that involve multi-output supervision, intermediate feature supervision, and the combination of multiple models within a single pipeline. This includes scenarios such as adversarial learning and teacher–student training.

KonfAI includes all the standard components of a deep learning pipeline, including data loading, training routines, logging, checkpointing, learning rate scheduling, and distributed execution. This section places emphasis on functionalities that are particularly important in medical imaging and often overlooked in other frameworks.

\paragraph{Data handling.}

KonfAI provides a flexible and extensible data management system adapted to the diversity of inputs in medical imaging workflows. All image data is accessed through a unified interface that supports both \texttt{SimpleITK}~\cite{mccormick2014itk} and \texttt{HDF5}, enabling direct use of standard formats including NIfTI, PNG, NRRD, MHA, and HDF5. This avoids format conversion and simplifies integration with existing datasets or institutional storage conventions.

Each dataset is structured into named groups that represent distinct elements of a case, such as an imaging modality or a label map. In file-based formats, groups correspond to individual files within a case directory. In HDF5 datasets, they are stored as internal subgroups for each entry. The configuration defines the role of each group as input, target, or auxiliary, and specifies how it should be preprocessed and patched. This structure ensures a clear and reproducible link between stored data, processing steps, and their usage within the model.

To manage large volumetric data and memory constraints, KonfAI includes native support for patch-based processing. During training, inference, and evaluation, volumes can be automatically split into overlapping patches. These patches may be 2D, 3D, or 2.5D, where the latter corresponds to stacks of adjacent slices along a single axis. This enables flexible trade-offs between spatial context, memory consumption, and architectural complexity. The 2.5D setting is especially useful when combining volumetric context with 2D convolutional backbones.

Patch reconstruction is handled transparently using configurable strategies such as mean fusion, cosine-weighted averaging, or custom aggregation. This ensures accurate volume reassembly while correctly managing overlaps, border regions, and potential missing data. Reconstruction is fully synchronized with the transformation and augmentation pipeline to preserve spatial consistency across all steps.

\paragraph{Test-Time Augmentation (TTA).}  
TTA is a widely used technique in medical AI to improve model robustness and generalization, particularly in tasks like segmentation and classification~\cite{wang2019aleatoric,moshkov2020test}. KonfAI implements TTA by applying a set of predefined transformations during inference and aggregating the resulting predictions through configurable strategies. These include \texttt{Mean}, \texttt{Median}, or any custom reduction function defined by the user. The process is fully controlled by the \texttt{reduction} parameter in the \texttt{Prediction.yml} file. In addition to improving prediction quality, this mechanism can be used to estimate uncertainty by measuring the variability across augmented outputs.

\paragraph{Model ensembling.}  
Model ensembling is widely used in medical image analysis to improve predictive accuracy, reduce variance, and increase robustness. It is a common component of high-performing solutions in segmentation challenges~\cite{ferreira2024we}.  

KonfAI enables ensembling by loading multiple trained models and combining their predictions during inference. Aggregation can be performed using uniform averaging or custom weighting, before applying the final post-processing steps. This behavior is fully controlled through the \texttt{combine} parameter in the \texttt{Prediction.yml} file, allowing ensemble inference without additional code.

Beyond performance gains, ensembling also enables the estimation of epistemic uncertainty by analyzing the variability across model predictions~\cite{kuestner2024predictive,lambert2024trustworthy}.

\paragraph{Multi-head supervision and intermediate feature access.}

Many medical imaging tasks require supervision across multiple outputs or targets. KonfAI supports multi-head architectures by allowing losses to be dynamically assigned to specific model heads and target groups, with custom weights and schedules defined in the configuration.

Recent deep learning approaches also rely increasingly on intermediate feature representations. These are essential in deep supervision~\cite{lee2015deeply}, multi-scale learning, feature distillation~\cite{zhang2021self}, and perceptual losses~\cite{johnson2016perceptual}, which compare activations from pre-trained networks rather than raw outputs.

Standard PyTorch models encapsulate internal layers within the \texttt{forward()} method, making intermediate features hard to access without altering the code. This often requires computing the full forward graph just to extract a single activation. KonfAI overcomes this by enforcing an explicit, hierarchical declaration of all submodules in the \texttt{\_\_init\_\_()} method. Each component, encoder, skip connection, attention block, head, can be referenced by name in the YAML configuration.

\newpage
For example, feature maps from a specific encoder block can be directly attached to a loss:

\begin{lstlisting}[caption={Example of perceptual loss configuration.}]
outputs_criterions:
  UNetBlock_0:DownConvBlock:
    targets_criterions:
      CT;MASK:
        criterions_loader:
          MAE:
            is_loss: true
            schedulers:
              Constant:
                nb_step: 0
                value: 1
\end{lstlisting}

In addition to custom architectures, predefined models from external libraries such as MONAI and Segmentation Models PyTorch~\cite{Iakubovskii2019} can also be imported and fully configured within the YAML files. This design promotes flexibility and allows users to rapidly build and experiment with a wide range of model architectures.

\paragraph{Flexible output handling.}

The \texttt{outputs\_dataset} section of the \texttt{Prediction.yml} file offers fine-grained control over how model outputs are post-processed and exported. Users can declaratively specify operations such as activation functions (softmax, argmax, binarization), data type casting, patch-based reconstruction, inverse transformations, and output formatting.

This mechanism ensures that predictions are directly usable for evaluation, visualization, or clinical integration, without requiring additional scripts or manual conversion steps. It supports seamless adaptation to downstream tools and guarantees consistency across experiments.

\paragraph{Evaluation and metrics reporting.}

KonfAI provides detailed evaluation outputs at both the individual case level and the dataset level, across training, validation, and testing splits. For each case, all configured metrics are computed per target group and saved in structured JSON files. This enables fine-grained performance analysis and facilitates targeted error diagnosis.

In parallel, summary statistics are computed for each metric, including mean, standard deviation, minimum, maximum, percentiles, and sample counts. These aggregated results offer a comprehensive view of model behavior across the dataset.

\section{Custom Module Integration}

KonfAI allows users to extend its functionality by implementing custom components, including models, loss functions, learning rate schedulers, data augmentations, transforms and more, without modifying the core library. New modules are implemented as Python classes that inherit from the appropriate abstract base class provided by the framework, such as \texttt{Loss}, \texttt{Scheduler}, \texttt{Augmentation}, \texttt{Transform}, or \texttt{Network}.

To integrate a new module, the user only needs to implement the required methods (typically \texttt{\_\_init\_\_} and \texttt{forward}), and place the file in an accessible location. Once defined, the class can be directly referenced in the YAML configuration using its name and file path. This mechanism makes it possible to inject custom logic into the training, prediction, or evaluation pipeline without altering the global structure of the framework.

For example, a custom loss \texttt{FocalLoss} can be implemented as:
\begin{lstlisting}[language=Python,caption={Minimal example of a custom loss in \texttt{CustomLoss.py}},label={lst:focal-loss}]
import torch
from konfai.criterions import Loss

class FocalLoss(Loss):
    def __init__(self, gamma: float, alpha: list[float], reduction: str="mean"):
        super().__init__()
        # Initialize parameters...

    def forward(self, input: torch.Tensor, target: list[torch.Tensor]) -> torch.Tensor:
        # Compute focal loss...
        return loss
\end{lstlisting}

Then reference it in the YAML config as:

\begin{lstlisting}[]
outputs_criterions:
  UNetBlock_0:Head:Softmax:
    targets_criterions:
      MASK:
        criterions_loader:
          CustomLoss:FocalLoss:
            is_loss: true
            gamma: 2.0
            alpha: [0.5, 3, 1.5]
            reduction: mean
\end{lstlisting}

\section{Conclusion}

KonfAI addresses a critical gap in the deep learning ecosystem for medical imaging by combining full declarative configuration with a modular and extensible architecture. Unlike existing frameworks that either impose rigid design patterns or couple configuration with code, KonfAI cleanly separates the what from the how, enabling researchers to specify complex pipelines through structured YAML files, without compromising flexibility or performance.

By supporting advanced workflows including patch-based training, 2.5D inference, test-time augmentation, model ensembling, and multi-output supervision, KonfAI addresses the specific needs of medical imaging. Its design promotes transparency, reproducibility, and fast experimentation, making it well suited for academic research.

The framework has been successfully applied to various real-world tasks, including segmentation, image registration, and image synthesis. It has also demonstrated robustness and scalability in competitive environments, contributing to top-ranking results in international MICCAI medical imaging challenges, including SynthRAD 2025 (Tasks 1 \& 2), TrackRAD 2025, CURVAS 2024, CURVAS PDACVI 2025, and PANTHER.
These results confirm KonfAI’s ability to manage complex, multimodal, and domain-shifted datasets while maintaining full reproducibility, modularity, and transparency across heterogeneous workflows.
Beyond these applications, KonfAI represents a step toward a broader vision: artificial intelligence as an active scientific agent. By providing a deterministic and transparent framework for experimentation, it enables the next stage of AI-driven discovery through language-driven deep learning experimentation, which we elaborate on in the following section.

\section{Perspectives — KonfAI-MCP: Toward Language-Driven Deep Learning Experimentation}

The frontier of large language models (LLMs) is shifting from understanding language to acting through it \cite{plaat2025agentic}.
Modern LLMs such as ChatGPT or Claude are no longer limited to generating text; they are evolving into agentic LLMs, capable of reasoning, planning, and interacting with computational tools \cite{matarazzo2025survey}.
Until now, LLMs have played a primarily descriptive role in deep learning research. They could explain how to train a network, suggest architectures, or generate some code; yet they remained observers rather than experimenters, unable to execute, verify, or refine what they proposed. Their reasoning stopped at the boundary of action.

While agentic LLMs mark a significant step toward actionable intelligence, their potential to conduct scientific experimentation remains largely unexplored. Looking ahead, we envision extending this paradigm toward deep learning experimentation itself, enabling systems where LLMs can directly specify, launch, refine, and reproduce model training workflows, from raw data to fully trained models addressing domain-specific tasks such as segmentation, registration, or image synthesis.
In such a framework, a single instruction such as “Train a model for lung tumor segmentation from this dataset” or “Reproduce and refine the experiment from this paper” could trigger a complete, auditable deep learning pipeline, encompassing data preprocessing, model configuration, optimization, evaluation, and iterative refinement.
Beyond simple execution, the agentic LLM becomes a proactive scientific collaborator—adapting existing methods, tuning hyperparameters, and proposing alternative architectures or data processing strategies based on observed outcomes and dataset characteristics. By closing the loop between conceptual design and empirical validation, it accelerates the integration of human ideas into executable and verifiable deep learning experiments, thus transforming language from a medium of description into a medium of empirical discovery.

Realizing such language-driven experimentation requires a substrate that connects probabilistic reasoning with deterministic execution.
Agentic LLMs fundamentally rely on external tools to act beyond text generation.
Early frameworks such as ReAct \cite{yao2022react}, HuggingGPT \cite{shen2023hugginggpt}, and AutoGen \cite{wu2024autogen} have demonstrated that effective agents must interleave reasoning and tool use to plan, retrieve, and execute complex computational tasks.
Building upon these advances, the Model Context Protocol (MCP) \cite{du2025mcp,anthropic2025tools} has emerged as an open interoperability standard that formalizes how LLMs can discover, invoke, and communicate with external tools through structured and typed interfaces.
MCP defines the syntax of agentic interaction, providing a secure and traceable communication layer between language models and computational environments, but it does not prescribe the semantics of execution within specific domains.
Consequently, even the most capable LLMs remain probabilistic, non-deterministic systems: their reasoning lacks grounding in executable, verifiable environments, thereby limiting reproducibility, auditability, and scientific reliability in applied workflows \cite{hou2025model,mohammadi2025evaluation}.

In this context, the effectiveness of an agentic LLM depends critically on the determinism, transparency, and semantic consistency of the tools it controls. High-quality tools must define strict input/output schemas, enforce validation, and return interpretable, context-rich responses under reproducible conditions \cite{anthropic2025tools}.

As a next step, we envision coupling KonfAI with MCP-enabled agents to materialize this vision. By serving as the deterministic, transparent, and semantically grounded substrate for deep learning experimentation, KonfAI could provide the execution layer through which LLMs transform natural language reasoning into verifiable and reproducible workflows. Such integration would establish a unified framework where language orchestrates the full lifecycle of deep learning research, from model specification to evaluation, laying the groundwork for reproducible, language-driven scientific discovery.

\section*{Acknowledgment}
This work was supported by the Brittany Region through its \textit{Allocations de Recherche Doctorale} program, and by the French National Research Agency (ANR) under the VATSop project (ANR-20-CE19-0015).

\bibliographystyle{plain} 
\bibliography{references}  

\newpage
\section*{Appendix A — Example Workspace} \label{Appendix:workspace-structure}

\begin{lstlisting}[caption={Structure of a KonfAI workspace after training, prediction, and evaluation of the UNet model.}]
Workspace/
  Checkpoints/
    UNet/
      2025_07_16_03_16_49.pt        # Trained model checkpoint (timestamped)
  Dataset/
    Patient_1/
      CT.mha                        # Input CT image
      MASK.mha                      # Ground truth segmentation
    Patient_2/
      CT.mha
      MASK.mha
  Evaluations/
    UNet/
      Evaluation.yml                # Evaluation configuration
      log.txt                       # Log of evaluation run
      Metric_EVALUATION.json        # Evaluation results on the test set
      Metric_TRAIN.json             # Evaluation results on the training set
  Predictions/
    UNet/
      Dataset/
        Patient_1/
          Seg.mha                   # Predicted segmentation for Patient_1
        Patient_2/
          Seg.mha
      log.txt                       # Log from prediction run
      Prediction.yml                # Prediction configuration
  Setups/
    UNet/
      Config_0.yml                  # Full training configuration snapshot
  Statistics/
    UNet/
      events.out.tfevents...        # TensorBoard logs (for training curves)
      log.txt                       # Training logs (text format)
  Config.yml                        # Main training configuration
  Evaluation.yml                    # Main evaluation configuration
  Prediction.yml                    # Main prediction configuration
  FocalLoss.py                      # Custom loss function (user-defined module)
\end{lstlisting}

\section*{Appendix B — Full Training Configuration (Config.yml)} \label{appendix:config-files}
\begin{lstlisting}[caption={Full training configuration (\texttt{Config.yml}) used in KonfAI to train a 2D UNet with patch-based learning, custom Focal Loss.},label={lst:config-full}]
Trainer:
  Model:
    classpath: segmentation.UNet.UNet     # Path to the model class
    UNet:
      Optimizer:                          # Optimizer configuration (AdamW)
        name: AdamW
        lr: 0.001                         # Initial learning rate
        betas: [0.9, 0.999]
        weight_decay: 0.01
      schedulers:                         # List of learning rate scheduler
        ReduceLROnPlateau:
          nb_step: 0
          factor: 0.1
          patience: 10
          threshold: 0.0001
      outputs_criterions:                 # Supervision attached to model outputs
        UNetBlock_0:Head:Softmax:         # Path to the output layer
          targets_criterions:
            MASK:                         # Supervised target group
              criterions_loader:
                FocalLoss:FocalLoss:      # User-defined Focal Loss
                  is_loss: true           # Loss used for training or just for tracking
                  schedulers:             # Defines how the loss weight evolves with iterations
                    Constant:
                      nb_step: 0
                      value: 1
                  gamma: 2.0
                  alpha: [0.5, 3, 1.5]
                  reduction: mean
      dim: 2                              # 2D model
      nb_batch_per_step: 1                # Number of batches before backward (accumulation)
      init_type: normal                   # Weight initialization method
      init_gain: 0.02
      BlockConfig:                        # Default block configuration
        kernel_size: 3
        stride: 1
        padding: 1
        bias: true
        activation: ReLU
        norm_mode: NONE
      channels: [1, 64, 128, 256, 512]    # Channel depth per stage
      nb_class: 2
      nb_conv_per_stage: 2
      downsample_mode: MAXPOOL
      upsample_mode: CONV_TRANSPOSE
      attention: false
      block_type: Conv
  Dataset:                                # Data loading, preprocessing, and patching
    groups_src:
      CT:                                 # Source group (e.g., CT image)
        groups_dest:
          CT:                             # Target group used internally
            transforms:                   # Preprocessing transforms
              Clip:
                min_value: -1000
                max_value: 1000
              Normalize:
                lazy: true
                channels: None
                min_value: -1
                max_value: 1
              ResampleToResolution:
                spacing: [1, 1, 1]
            patch_transforms: None
            is_input: true                # true if this input is fed to the model
      MASK:                               # Source group for mask (segmentation target)
        groups_dest:
          MASK:
            transforms:
              ResampleToResolution:
                spacing: [1, 1, 1]
            patch_transforms: None
            is_input: false
    augmentations:                        # Data augmentation pipeline
      DataAugmentation_0:
        data_augmentations:
          Flip:                           # Random flip with axis probabilities
            prob: 1.0
            f_prob: [0, 0.5, 0.5]
        nb: 1                             # Number of augmentations to apply per sample
    Patch:                                # Patch extraction parameters
      patch_size: [1, 224, 288]
      overlap: None
      extend_slice: 2                     # Number of adjacent slices to include (2.5D mode)
      pad_value: -1
    subset: None                          # Define a subset of the dataset (file or list of indices)
    shuffle: true
    dataset_filenames:                    # Path to dataset files
      - ./Dataset/:a:mha # In .mha format here, supports any format handled by SimpleITK or HDF5
    use_cache: true                       # Whether to load all data into RAM before training
    batch_size: 5
    validation: None                      # Validation file or split ratio (e.g. 0.2).
    inline_augmentations: false
  train_name: UNet                        # Name of the experiment
  manual_seed: 32                         # Reproducibility seed
  epochs: 100
  it_validation: None
  autocast: false                         # Use AMP (automatic mixed precision)
  gradient_checkpoints: None              # Enable gradient checkpointing to save memory
  gpu_checkpoints: None                   # Change GPU from this layer onward
  ema_decay: 0                            # Exponential Moving Average (EMA) decay rate
  data_log:                               # Data to visualize during training
    - CT/IMAGES/5
    - MASK/IMAGES/5
    - UNetBlock_0:Head:Argmax/IMAGES/5
  save_checkpoint_mode: BEST              # Save only the best checkpoint (ALL/BEST)
  EarlyStopping:                          # Early stopping criteria
    monitor: None
    patience: 10
    min_delta: 0.0
    mode: min
\end{lstlisting}

\begin{lstlisting}[caption={Prediction configuration (\texttt{Prediction.yml}) used in KonfAI to apply a trained 2D UNet with patch-based inference and post-processing.},label={lst:config-prediction}]
Predictor:
  Model:
    classpath:  segmentation.UNet.UNet   # Path to the model definition
    UNet:
      outputs_criterions: None          # Metric can be define for tracking
      BlockConfig:                      # Default block configuration
        kernel_size: 3
        stride: 1
        padding: 1
        bias: true
        activation: ReLU
        norm_mode: NONE
      dim: 2                            # 2D model
      channels: [1, 64, 128, 256, 512]  # Channel depth per stage
      nb_class: 2
      nb_conv_per_stage: 2
      downsample_mode: MAXPOOL
      upsample_mode: CONV_TRANSPOSE
      attention: false
      block_type: Conv
  Dataset:                              # Input dataset for inference
    groups_src:
      CT:                               # Source group (e.g., CT image)
        groups_dest:
          CT:                           # Target group used internally
            transforms:                 # Preprocessing transforms
              Clip:
                min_value: -1000
                max_value: 1000
                save_clip_min: false
                save_clip_max: false
              Normalize:
                lazy: true
                channels: None
                min_value: -1
                max_value: 1
              ResampleToResolution:
                spacing: [1, 1, 1]
            patch_transforms: None
            is_input: true
    augmentations:                      # Test-time augmentations
      DataAugmentation_0:
        data_augmentations:
          Flip:
            prob: 1
            f_prob: [0, 1, 0]
        nb: 1
      DataAugmentation_1:
        data_augmentations:
          Flip:
            prob: 1
            f_prob: [0, 0, 1]
        nb: 1
    Patch:                              # Patch-based inference config
      patch_size: [1, 512, 512]
      overlap: None
      extend_slice: 2
      pad_value: -1
    subset: None
    dataset_filenames:
      - ./Dataset/:a:mha
    batch_size: 32
  outputs_dataset:                      # How predictions are saved
    UNetBlock_0:Head:Softmax:           # Path to predicted tensor
      OutputDataset:
        name_class: OutSameAsGroupDataset
        before_reduction_transforms: None
        after_reduction_transforms:     # Apply argmax over softmax
          Argmax:
            dim: 0
        final_transforms:               # Cast to final type
          TensorCast:
            dtype: uint8
        same_as_group: MASK:MASK        # Save result in same shape as MASK
        dataset_filename: ./Dataset:mha # Output file format
        group: seg
        patch_combine: None             # Default: average overlapping regions
        reduction: Mean
        inverse_transform: true         # Apply inverse of original transform
  combine: Mean                         # Combine augmented predictions
  train_name: UNet                      # Name of model directory
  manual_seed: None
  gpu_checkpoints: None
  autocast: false                       # Use FP32 inference
  data_log: None
\end{lstlisting}

\begin{lstlisting}[caption={Evaluation configuration (\texttt{Evaluation.yml}) used in KonfAI to compute metrics between predictions and ground truth. A summary JSON is written per sample and globally.},label={lst:evaluation-config}]
Evaluator:
  metrics:                              # Metrics to compute for each target group
    MASK:                               # Target name for evaluation
      targets_criterions:
        MASK_Pred:                      # Prediction to compare with target
          criterions_loader:
            Dice:                       # Metric used (e.g., Dice coefficient)
              smooth: 1e-06
  Dataset:                              # Dataset containing ground truth and prediction
    groups_src:
      MASK:                             # Ground truth segmentation
        groups_dest:
          MASK:
            transforms: None
      MASK_Pred:                        # Predicted segmentation
        groups_dest:
          MASK_Pred:
            transforms: None
    subset: None
    dataset_filenames:
      - ./Dataset:a:mha                 # Path to ground truth data
      - ./Predictions/UNet/Dataset/:i:mha    # Path to prediction data
    validation: None                    # No validation subset used
  train_name: UNet                      # Name of the experiment
\end{lstlisting}

\end{document}